\documentclass[10pt,twocolumn,letterpaper]{article}

\usepackage{cvpr}              

\usepackage[dvipsnames]{xcolor}

\definecolor{cvprblue}{rgb}{0.21,0.49,0.74}
\usepackage[pagebackref,breaklinks,colorlinks,citecolor=cvprblue]{hyperref}

\def\paperID{14651} 
\def\confName{CVPR}
\def\confYear{2024}

\title{Novel Class Discovery for Ultra-Fine-Grained Visual Categorization}

\author{Yu Liu$^1$, Yaqi Cai$^1$, Qi Jia$^{1}$\thanks{corresponding author}, Binglin Qiu$^1$, Weimin Wang$^1$, Nan Pu$^2$\\
$^1$International School of Information Science and Engineering, Dalian University of Technology \\
$^2$Department of Information Engineering and Computer Science, University of Trento\\
{\tt\small \{liuyu8824,jiaqi,wangweimin\}@dlut.edu.cn, \{caiyaqi1998,m1andy\}@mail.dlut.edu.cn, } \\
{\tt\small nan.pu@unitn.it}
}
\begin{document}
\maketitle
\begin{abstract}

Ultra-fine-grained visual categorization (Ultra-FGVC) aims at distinguishing highly similar sub-categories within fine-grained objects, such as different soybean cultivars. 
Compared to traditional fine-grained visual categorization, Ultra-FGVC encounters more hurdles due to the small inter-class and large intra-class variation. 
Given these challenges, relying on human annotation for Ultra-FGVC is impractical. 
To this end, our work introduces a novel task termed Ultra-Fine-Grained Novel Class Discovery (UFG-NCD), which leverages partially annotated data to identify new categories of unlabeled images for Ultra-FGVC. 
To tackle this problem, we devise a Region-Aligned Proxy Learning (RAPL) framework, which comprises a Channel-wise Region Alignment (CRA) module and a Semi-Supervised Proxy Learning (SemiPL) strategy. 
The CRA module is designed to extract and utilize discriminative features from local regions, facilitating knowledge transfer from labeled to unlabeled classes. 
Furthermore, SemiPL strengthens representation learning and knowledge transfer with proxy-guided supervised learning and proxy-guided contrastive learning. 
Such techniques leverage class distribution information in the embedding space, improving the mining of subtle differences between labeled and unlabeled ultra-fine-grained classes.
Extensive experiments demonstrate that RAPL significantly outperforms baselines across various datasets, indicating its effectiveness in handling the challenges of UFG-NCD. Code is available at \href{https://github.com/SSDUT-Caiyq/UFG-NCD}{https://github.com/SSDUT-Caiyq/UFG-NCD.}
\end{abstract}

\section{Introduction}
\label{sec:intro}

\begin{figure}[t]
  \centering
   \includegraphics[width=1\linewidth]{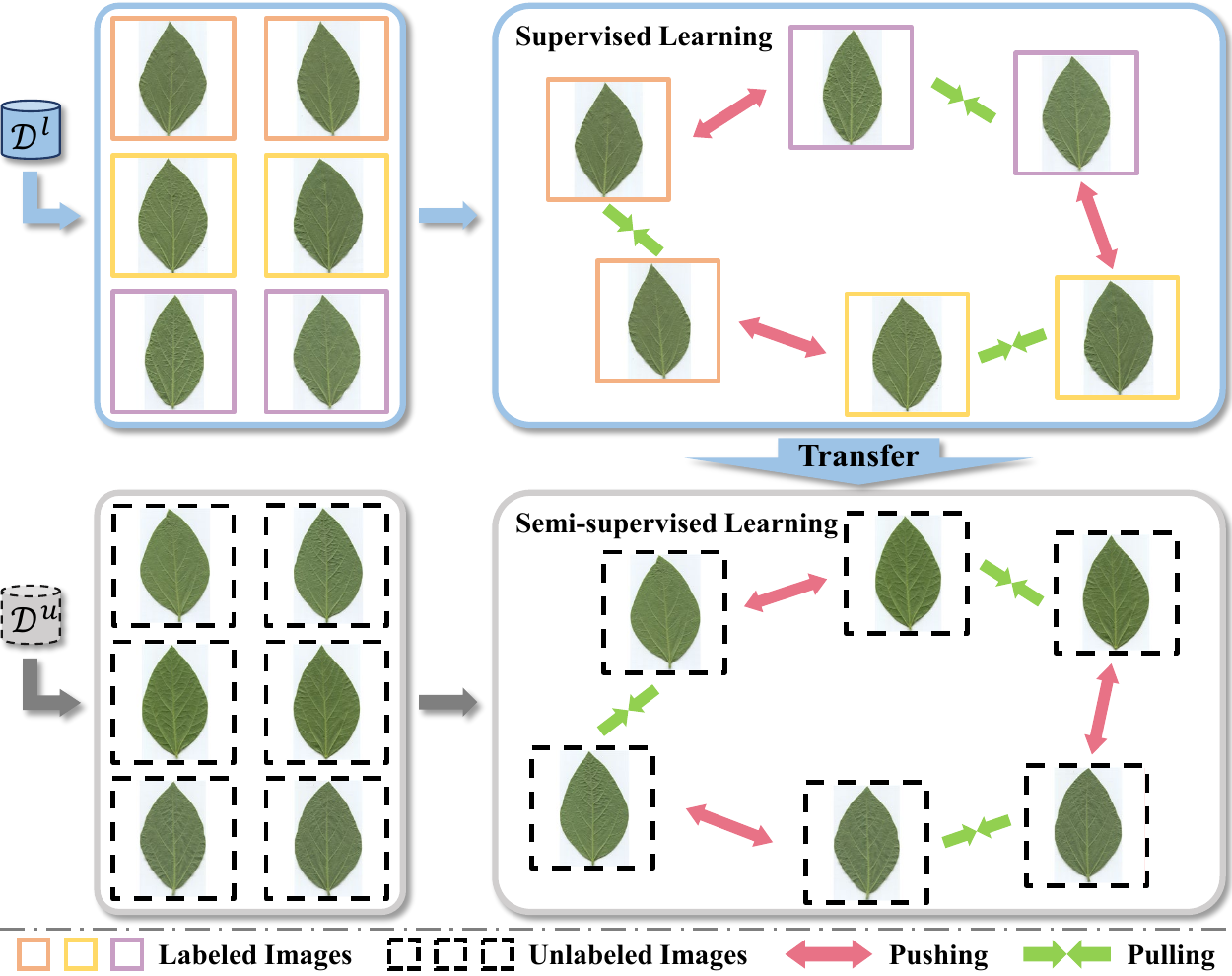}
   \caption{Visual conception of UFG-NCD, where $D^l,D^u$ are labeled and unlabeled data splits. 
   We propose to exploit novel class discovery for partially annotated UFG images. 
   The core idea is to learn prior knowledge from labeled images via supervised learning, subsequently extending the knowledge to unlabeled images via a semi-supervised framework. It facilitates knowledge transfer and representation learning across ultra-fine-grained classes.
   }
   \label{fig:ufg-ncd}
\end{figure}

``There are no two identical leaves in the world'', Leibniz once said. 
However, it is trivial and troublesome to identify the subtle differences given visually similar leaf images, even for senior botanists. 
Traditional fine-grained visual categorization (FGVC)~\cite{fgvc01-embedding} has enabled the differentiation of subordinate categories within a broader class, such as bird species~\cite{cub}, by focusing on distinct region features like head and wing. 
Yet, FGVC methods are still limited when it comes to dissecting these categories further into sub-categories.
To this end, ultra-fine-grained visual categorization (Ultra-FGVC)~\cite{ultra-dataset21,ultra-ffvt21,ultra-maskcov21,ultra-spare22,ultra-ssfe23,ultra-mix-vit23,ultra-cle-vit23} is researched at recognizing sub-categories of ultra-fine-grained (UFG) images from a specific fine-grained category, which attracts the attention of researchers due to its significant potential in real-world applications like precision agriculture. 
In~\cref{fig:ufg-ncd}, for example, each row of leaf images corresponds to a distinct soybean cultivar, but the differences between these UFG classes are hardly able to be defined and perceived by humans~\cite {ultra-dataset21}. 
One can expect that, Ultra-FGVC suffers more from small inter-class and large intra-class variation inherent to FGVC. 
Hence, it is impractical to fully annotate UFG image instances for a supervised learning paradigm. 

In recent years, an emergent task termed novel class discovery (NCD)
~\cite{ncd-dtc,ncd-rankstat,ncd-uno,ncd-comex,ncd-iic,ncd-rkd}
gathers more attention due to its ability to discover novel categories from a pool of unlabeled instances in a semi-supervised fashion.
Note that, NCD task has access to some labeled classes disjoint from the unlabeled ones.
The success of NCD relies on conceptual knowledge transfer from labeled to unlabeled images. 
It means that the model pre-trained on labeled data needs to learn discriminative representation suitable for clustering unlabeled samples. 
A line of early works is dedicated to modeling the relation of unlabeled images via clustering objectives~\cite{ncd-dtc} or soft class predictions~\cite{ncd-rankstat,ncd-uno}.
However, these approaches neglect the semantic relations between labeled and unlabeled classes, thereby hindering the transfer learning of discriminative representation. 
To this end, recent works~\cite{ncd-iic,ncd-rkd} 
focus more on capturing instance-level constraints in the feature or prediction space. 
Despite the empirical effectiveness of such methods on generic and fine-grained NCD benchmarks, they are limited by learning imperceptible clues from unlabeled UFG images.

In this work, we exploit novel class discovery for 
addressing Ultra-FGVC, with no need to fully annotate all the images.
This new task setting termed UFG-NCD has not been studied in the research community yet. 
To tackle this problem, we propose a simple yet effective framework called region-aligned proxy learning (RAPL), 
which is tailored specifically for UFG-NCD.
First of all, we develop a region-aligned module on the feature maps, which groups different channels and forces each group to discover discriminative features in a local region. 
In addition, inspired by proxy-based methods~\cite{dml-proxygml,dml-proxyanchor,dml-proymulti} in deep metric learning,
we present a novel semi-supervised proxy learning (SemiPL) framework, modeling global structure in feature embedding space and facilitating representation learning by using the relation between class-specified proxies and data samples. 
In this way, the labeled and unlabeled data can be optimized jointly by proxy-guided supervised learning (PSL) and proxy-guided contrastive learning (PCL), respectively.
Last but not least, our RAPL is also applicable to generalized category discovery (GCD) setting~\cite{gcd-gcd,gcd-dynamic}, where the unlabeled data include both labeled and unlabeled categories.
Our contributions are summarized as follows:
\begin{itemize}
    \item [$\bullet$]
    We are the first to explore how to categorize unlabeled UFG images with labeled ones of disjoint classes, resulting in a new and practical task termed as UFG-NCD.
    \item [$\bullet$]
    We propose a channel-wise region alignment module to 
    reveal subtle differences within a local region, which promotes the learning and transfer of local representation.
    \item [$\bullet$]
    We introduce a novel two-stage semi-supervised proxy learning framework for UFG-NCD, which models the relationship between instances and class proxies in embedding space, and learns from unlabeled images with global distribution information.
    \item [$\bullet$]
    We introduce five UFG-NCD datasets, where extensive experiments demonstrate our method achieves significant improvements over previous NCD methods.
\end{itemize}

\section{Related Work}
\label{sec:relatedwork}
\begin{figure*}[t]
  \centering
   \includegraphics[width=1\linewidth]{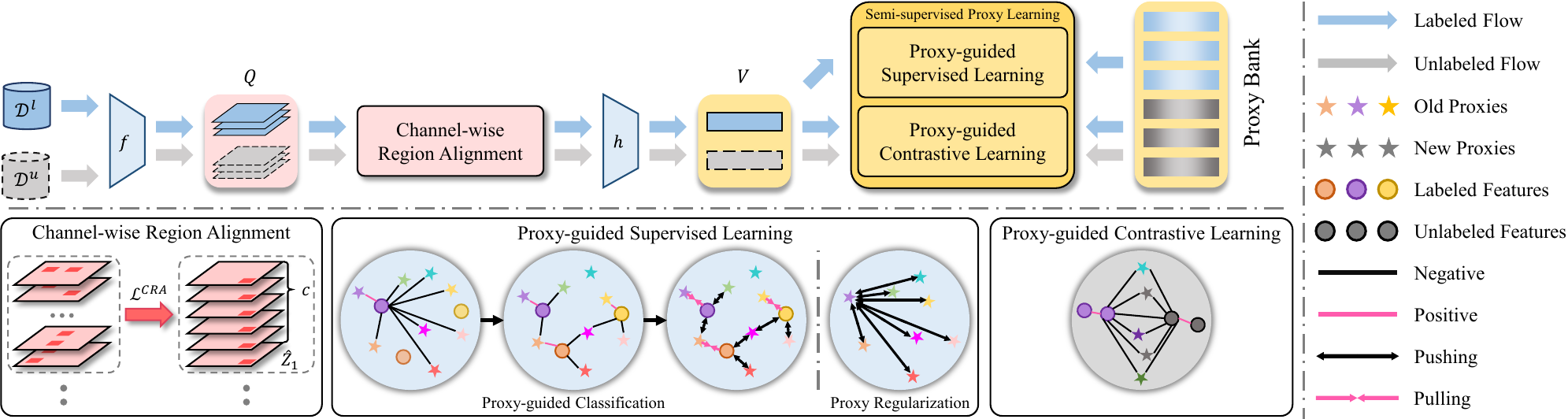}
   \caption{Overview of our Region-Aligned Proxy Learning (RAPL) framework. We first group and align extracted feature maps in the set $Q$ with regions via Channel-wise Region Alignment. 
   Then, we generate global representation $v$ through global average pooling and MLP projection $h$, and lastly learn and transfer knowledge with the proxy bank using semi-supervised proxy learning (SemiPL).
   }
   \label{fig:overview}
\end{figure*}

\subsection{Ultra-Fine-Grained Visual Categorization} 
Yu et al.~\cite{ultra-dataset21} pioneered the formulation of the ultra-fine-grained visual categorization (Ultra-FGVC) task, with the objective of discerning sub-categories within specific fine-grained categories. 
Concurrently, they introduced two ultra-fine-grained (UFG) datasets focused on leaves, namely SoyAgeing and SoyGene.
The collected images of the same sub-category correspond to specific seeds obtained from the genetic resource repository. 
To address the challenges posed by the small inter-class and large intra-class variations in Ultra-FGVC, existing approaches typically fall into three categories: local-based~\cite{ultra-maskcov21, ultra-spare22}, feature fusion-based~\cite{ultra-ffvt21, ultra-ssfe23, ultra-mix-vit23}, and contrastive learning-based~\cite{ultra-cle-vit23} methods. First, akin to conventional fine-grained classification~\cite{fgvc-mcloss, fgvc-navigate, fgvc-ssllocal}, local-based approaches focus on learning local embeddings by predicting masked or erased regions within images. For example, MaskCOV~\cite{ultra-maskcov21} employed a covariance matrix to capture mutual information between each quarter of randomly masked and shuffled image patches. Second, feature fusion-based approaches enhance local and global discrimination by fusing important tokens, drawing inspiration from the attention mechanism in vision transformers. For instance, FFVT~\cite{ultra-ffvt21} introduced a novel mutual attention weight selection module that chose the token most similar to each anchor token. 
Third, motivated by the success of contrastive learning in self-supervised learning, CLE-ViT~\cite{ultra-cle-vit23} generated positive samples through randomly masked and shuffled image parts, thereby mitigating the challenges of class variation.
Notably, the effectiveness of these approaches relies on fully labeled data, which is challenging and expensive to obtain for UFG images in practical scenarios.

In contrast to these methods, our work enables the categorization of unlabeled UFG images in a semi-supervised learning fashion, leveraging partially labeled data.


\subsection{Novel Class Discovery} 
Different from conventional semi-supervised learning~\cite{semisup-1,semisup-2,semisup-3}, novel class discovery (NCD) infers unlabeled images from related but disjoint classes with labeled ones. 
In order to take advantage of labeled data, most of the approaches~\cite{ncd-dtc,ncd-rankstat,ncd-joint,ncd-dualrank,ncd-openmix,ncd-ncl,ncd-restune,ncd-uno,ncd-comex,ncd-iic,ncd-rkd} conduct two training stages. 
In the first stage, the model is pre-trained with the labeled data for a standard supervised learning objective; for the second stage, the knowledge learned from the model is 
transferred to group the unlabeled data based on elaborately designed clustering or self-supervised objectives.
Specifically, early work~\cite{ncd-rankstat,ncd-dualrank} modeled the binary relationship between samples based on the similarity of the top-k feature dimensions. 
UNO~\cite{ncd-uno} built a connection of images and classes by the Sinkhorn-Knopp algorithm~\cite{sinkhorn}. 
Meanwhile, a line of work~\cite{ncd-ncl,ncd-openmix} employed richer inter-sample relations with nearest neighbor in feature space. 
However, these instance-level or feature-level methods neglect rich inter-class relations, 
leading to compact class distributions that hinder the representation learning of new categories. 
Thus, recent works~\cite{ncd-iic,ncd-rkd} were dedicated to leveraging inter-class similarity information. 
OpenLDN~\cite{ncd-openldn} further explored pairwise similarity relations to regularize feature distributions of labeled and unlabeled data. 
Nevertheless, these methods are much challenged by small inter-class variance in UFG sub-categories. In this work, we proposed a novel semi-supervised proxy learning framework to explicitly learn inter-class discriminative knowledge through class-specified proxies in feature space. 

\section{Proposed Method}

\subsection{Problem Formulation}
Denote $\mathcal{D}$ as a UFG dataset, including a labeled split $\mathcal{D}^l = \{(x_i^l,y_i^l)\} ^{N^l}_{i=1}$ and an unlabeled split $\mathcal{D}^u = \{(x_i^u,y_i^u)\} ^{N^u}_{i=1}$, where $y$ is the class label of image $x$, and $N^l$ and $N^u$ are the number of instances in two data splits. 
It is noteworthy that, following NCD, we assume labeled images from old classes, \ie, $y^l \in \mathbb{R}^{C^l}$, while unlabeled ones are from disjoint new classes to be classified, \ie, $y^u \in \mathbb{R}^{C^u}$, and $C^l \cap C^u = \emptyset$. 
During the training phase, we assume labels $y^l$ are available, whereas $y^u$ remain unknown. 
Our objective is to classify the unlabeled split based on the labeled one.
Consistent with most prior research~\cite{ncd-rankstat,ncd-uno}, the number of new classes $C^u$ is assumed to be a known priori. 


\subsection{Overview}
As depicted in~\cref{fig:overview}, we propose a novel region-aligned proxy learning (RAPL) framework that consists of three main designs for UFG-NCD.
First of all, we develop a novel channel-wise region alignment (CRA) module, which aims at explicitly learning local fine-grained representation in distinct regions with their corresponding groups of feature maps.
After locating regional patterns via CRA, we introduce a semi-supervised proxy learning (SemiPL) approach, integrating proxy-guided supervised learning (PSL) and proxy-guided contrastive learning (PCL) simultaneously. 
PSL learns a global representation and regularizes the distribution of embedding space via proxy-guided classification and proxy regularization.
Besides, we adopt PCL to facilitate representation learning on unlabeled data with global distribution information of class proxies. 

\subsection{Channel-wise Region Alignment}
The distinctions among ultra-fine-grained classes become less perceptible compared to classes in conventional FGVC. 
Consequently, categorizing these classes heavily depends on local discriminative features.
For instance, MaskCOV~\cite{ultra-maskcov21} and SPARE~\cite{ultra-spare22} attend local information via semantic context constraints across randomly masked or erased image patches. 
These methods mainly model patch-to-patch relations in a certain image for local representation learning, whereas such relations are unable to be shared and transferred from labeled samples to unlabeled ones.
Thus, we propose a channel-wise region alignment (CRA) module to leverage local representation for distinctive feature learning in UFG-NCD.
As illustrated in the bottom-left subplot of Fig.~\ref{fig:overview}, we explicitly organize and associate feature channels with specific regions to extract subtle visual cues within each region. The local features learned in CRA are discriminative for all classes, advancing the transfer of knowledge from labeled classes to unlabeled ones. 

Specifically, given an input image $x_i$, the feature maps extracted by an image encoder $f$ is denoted as $Z_i = \{z_i^1,..,z_i^{D} \} \in \mathbb{R}^{D\times W\times H}$.  Here, $Z_i$ consists of $D$ channels of feature matrix $z$ with dimensions of height $H$ and width $W$, where each dimension of $z$ encodes a region representation of the image.
In order to assign distinct weights to each local region,
we divide $D$ feature channels into $H * W$ groups, denoted as $\{\hat{Z}^0_i,\dots,\hat{Z}^{H*W-1}_i\}$, 
Each group is responsible for encoding a specific region representation. In this way, each group of feature channels is defined as 
\begin{equation}
    \hat{Z}^j_i = \{z^{k}_i  \in \mathbb{R}^{H \times W} | k \in [j\times c + 1,..,j \times c + c] \},
\end{equation}
where each group comprises $c=\frac{D}{H * W}$ channels.
Subsequently, a region label $y_i^k = j$ is assigned to each channel $z^k_i$ in the group $\hat Z^j_i$, explicitly linking feature channels within each group to a specific region. 
As a result, we enforce each feature channel to concentrate on its corresponding region through a loss function denoted as $\mathcal{L}^{CRA}$, calculated by applying standard cross-entropy on each flattened feature matrix, \ie, $\text{flatten}(z_i)\in \mathbb{R}^{H * W}$:
\begin{equation}
    \mathcal{L}^{CRA} = \frac{1}{D}\sum^{D}_{k=1} - y_i^k \text{log}(\text{flatten}(z^k_i)).
    \label{eq:cra}
\end{equation}


After learning regional patterns via CRA, it is crucial to 
combine them into a global representation for the final classification.
As shown in Fig.~\ref{fig:overview},
we achieve a global representation via global average pooling and MLP projection on the flattened feature maps, which are denoted as set of feature vectors $V= h(\text{flatten}(Q)) \in \mathbb{R}^{N \times D}$, where $h(\cdot) = \text{MLP}(\text{GAP}(\cdot))$, $Q = \{Z_1,..Z_N\}$.

\subsection{Semi-supervised Proxy Learning}
In the context of Ultra-FGVC, the primary challenge is to capture subtle differences among similar samples that prove challenging to discern in feature space. The use of a linear classifier is a common approach to optimizing feature representation. However, this method is susceptible to overfitting on labeled classes, leading to the encoding of insufficient and less discriminative knowledge when faced with unlabeled classes~\cite{gcd-gcd}. 
Thus, we propose a novel semi-supervised proxy learning (SemiPL) approach for UFG-NCD, which encodes discrimination and distribution of labeled instances via proxy-guided supervised learning, and further 
transfers suitable knowledge from labeled images to unlabeled ones by proxy-guided contrastive learning. 

\noindent \textbf{Proxy-guided Supervised Learning (PSL).}
Rather than assigning a specific pseudo-label to each instance, we introduce class-level proxies to establish relations between instances and classes, and further learn representation and regularize distribution in feature space via proxy-guided classification and proxy regularization, respectively, drawing inspiration from works such as~\cite{dml-proxygml,dml-proxyanchor,dml-proymulti}. 

\begin{figure}[t]
  \centering
   \includegraphics[width=0.9\linewidth]{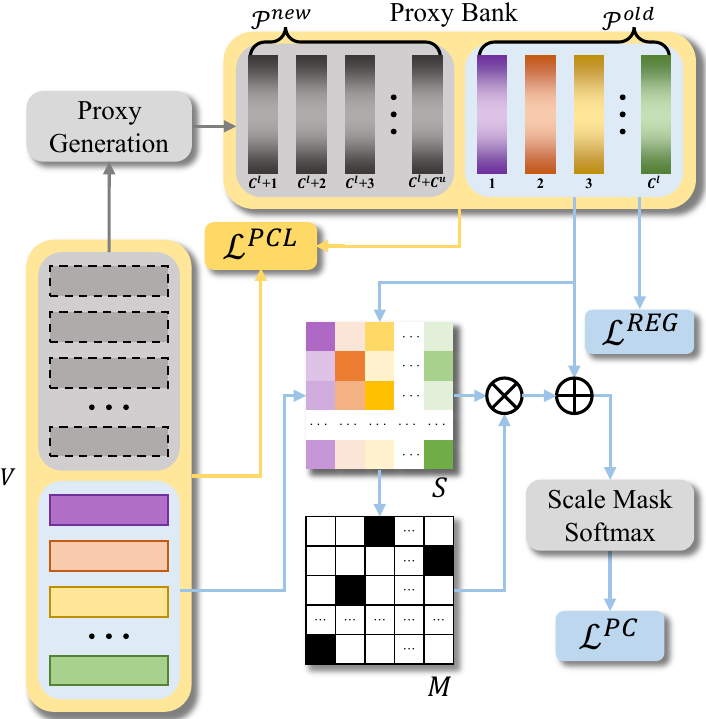}
   \caption{Illustration of proposed SemiPL. We first introduce a supervised paradigm, indicated by blue lines, which includes $\mathcal{L}^{PC}$ and $\mathcal{L}^{REG}$, representing the loss of proxy-guided classification and proxy regularization, respectively. Then, we propose PCL with global distribution guidance of $\mathcal{P}^{old}$ and $\mathcal{P}^{new}$ to learn and discover new classes from unlabeled data, indicated in yellow lines.}
   \label{fig:proxylearning}
\end{figure}

For proxy-guided classification, we first generate classification probabilities by comparing each visual feature 
with its $\{k+1\}$-nearest proxies, instead of learning a linear classifier. 
We aim to lower the probabilities of negative proxies by loss constraints during training.
Initially, we construct a set of learnable proxies $\mathcal{P}^{old}$ according to labeled classes in feature space, which is denoted as $\mathcal{P}^{old} =\{(p^l_i,y^l_i)|i\in [1,..,C^l], p^l_i \in \mathbb{R}^D, y^l_i=i\}$. 
Subsequently, given features $V^l=\{v^l_1,..,v^l_{N^l}\}$ of labeled images, we compute the classification probabilities with cosine similarity:
\begin{equation}
    S = \{ s_{ij}=(v^{l}_i)^{\text{T}}p^l_j | i\in [1,..,N^l], j\in [1,..,C^l]\},
    \label{eq:similarity_sample_to_proxy}
\end{equation}
where $s_{ij}$ indicates the similarity between the $i$-th instance $v^{l}_i$ and the $j$-th proxy $p^l_j$. In the leftmost circle of the proxy-guided classification subplot in~\cref{fig:overview}, let us take the feature labeled by the purple circle as an example, and the colorful stars represent the proxies. The pink line indicates the only positive relation between the feature and the proxy to which it belongs, while the black lines represent the negative relations between the feature and other proxies. 



In conventional classification scenarios, $s_{ij}$ should be significantly higher than others if the $i$-th instance belongs to the $j$-th class proxy. However, for Ultra-FGVC, each instance may have several negative relations with high probabilities across different classes due to small intra-class variation. 
To lower the high probabilities in negative relations at training phase, we select the top $k$ negative relations using a mask matrix $M$ with the same size as $S$. 
In particular, the selected probabilities in $M$ are marked with $1$, while others are marked with $0$. Formally, each element in the mask $M \in \mathbb{R}^{N^l \times C^l}$ is defined by:
\begin{equation}
   m_{ij} = 
   \begin{cases}
       1, &\text{if~} s_{ij} \in \text{top-}k(S_{i} \backslash s_{ig}),~g=y_{i} \\
       0, &\text{else},
   \end{cases}
   \label{eq:mask_m}
\end{equation}
where $S_{i}=[s_{i1},..,s_{iC^l}]$ represents the similarity between the $i$-th instance and all old proxies, 
and $s_{ig}$ is the similarity between $i$-th instance and proxy of the correct class $y_i$ to which the instance belongs, which is the probability of positive relation.
Then, the class prediction improved by the mask becomes
\begin{equation}
    \overline{Y} = S \odot M + S^{pos},
    \label{eq:select_similarity}
\end{equation}
where $S^{pos} = \{s_{ig} \in S|g=y_{i}\}$ is the set only including probabilities of positive relations. 
As shown in the second circle of the middle bottom subfigure in~\cref{fig:overview}, each feature has one positive relation and $k$ negative relations with corresponding proxies.
Note that, $\overline{Y}$ is sparse due to a limited $k$, which leads to an inflated denominator in conventional softmax. 
Thus, we employ an indicator function $\mathbbm{1}(\cdot)$ to eliminate the effect of redundant zero elements in $\overline{Y}$, and thereby the final prediction is 
\begin{equation}
    \tilde{Y}_{ij} = \frac{ \mathbbm{1}(\overline{Y}_{ij} \neq 0) \text{exp}(\overline{Y}_{ij}) }{ \sum^{C^l}_{g=1} \mathbbm{1}(\overline{Y}_{ig} \neq 0) \text{exp}(\overline{Y}_{ig})},
    \label{eq:sacle_mask_softmax}
\end{equation}
where $\mathbbm{1}(\cdot)$ is 1 when $\overline{Y}_{ij} \neq 0$; otherwise is 0. 
Based on the prediction probabilities, we can define the loss of proxy-guided classification as
\begin{equation}
    \mathcal{L}^{PC} = -\frac{1}{N^l} \sum^{N^l}_{i = 1} \sum^{C^l}_{j=1} \mathbbm{1}(y_i=j) \text{log}(\tilde{Y}_{ij}).
    \label{eq:pc}
\end{equation}

As shown in the third circle at the bottom of Fig.~\ref{fig:overview}, the positive relation between the feature and corresponding proxy are pulled close while negative relations are pushed away during training.

As each proxy represents the center of a class, we expect the difference between proxies to be as large as possible. 
Furthermore, we introduce a regularization constraint to make the learned proxies as discriminative as possible.
We compute the similarity between old proxies by
\begin{equation}
    S^p = \{ s^p_{ij}=(p^l_i)^{\text{T}}p^l_j | i,j\in [1,..,C^l]\}.
    \label{eq:similarity_proxy_to_proxy}
\end{equation}

Afterward, the proxy regularization is defined as
\begin{equation}
    \mathcal{L}^{REG} = -\frac{1}{C^l} \sum^{C^l}_{i=1} \sum^{C^l}_{j=1} \mathbbm{1}(i=j) \text{log}(\text{softmax}(S^p_{ij})).
    \label{eq:reg}
\end{equation}

It is worth noting that $\mathcal{L}^{REG}$ is conducted in feature space, which regularizes class distribution via similarity between class-specified proxies. The fourth circle at the bottom of Fig.~\ref{fig:overview} illustrates the proxies are pulled away properly thanks to imposing proxy regularization.

\noindent\textbf{Proxy-guided Contrastive Learning (PCL).}
Considering the knowledge transfer and learning of unlabeled data,
it is difficult to define the positive relations between unlabeled instances.
Like recent work~\cite{ncd-uno,ncd-ncl}, we take advantage of contrastive learning to learn instance-wise representation with two augmented views of each sample.

Before developing proxy-guided contrastive learning (PCL), we firstly capture global structure information of unlabeled classes by generating a set of proxies in feature space accordingly, which is marked as $\mathcal{P}^{new} =\{(p^u_i,y^u_i)|i\in [C^l + 1,.., C^l + C^u], p^u_i \in \mathbb{R}^D, y^u_i=i\}$. 
As shown in yellow lines of~\cref{fig:proxylearning}, we employ the guidance of both $\mathcal{P}^{old}$ and $\mathcal{P}^{new}$ to mine discriminative features in contrastive learning.
Thus, the input vectors $\overline{V}$ of contrastive learning become a combination of $v, v'$, and $\mathcal{P}^{old}$, $\mathcal{P}^{new}$, where the $v,v'$ are the feature vectors extracted from two views of images. 
Note that, the positive vector for each proxy is itself in constrastive learning.
The loss of PCL is written as
\begin{equation}
    \mathcal{L}^{PCL} = - \sum^{|\overline{V}|}_{i=1}\text{log}\frac{\text{exp}(\overline{v}_i\cdot \overline v_i' / \tau)}{\sum^{|\overline{V}|}_{j=1} \mathbbm{1}(i\neq j)\text{exp}(\overline v_i\cdot \overline v_j \tau)},
    \label{eq:pcl}
\end{equation}
where $\overline{v}_i, \overline{v}'_i$ is a feature vector and its positive vector in contrastive learning,
$\tau$ is the temperature, and $|\overline{V}|$ is number of feature vectors in $\overline{V}$. 
Proxy-guided contrastive loss learns instance-level representation based on region-aligned feature maps and separate unlabeled data points in the embedding space.
Note that, PCL achieves knowledge transfer without introducing any inaccurate positive relation, which optimizes feature distribution with the guidance of proxies. 

\begin{table*}[t]
\centering
  \begin{adjustbox}{width=1\linewidth}
    \begin{tabular}{cccccccccccccccc}
      \toprule
      \multirow{2}{*}{Method} & \multicolumn{3}{c}{SoyAgeing-R1} & \multicolumn{3}{c}{SoyAgeing-R3} & \multicolumn{3}{c}{SoyAgeing-R4} & \multicolumn{3}{c}{SoyAgeing-R5} & \multicolumn{3}{c}{SoyAgeing-R6} \\
        & All & Old & New & All & Old & New & All & Old & New & All & Old & New & All & Old & New \\
      \midrule
      RankStats IL~\cite{ncd-rankstat} & 42.12 & 71.92 & 12.32 & 42.53 & 71.31 & 13.74 & 40.10 & \underline{69.70} & 10.51 & 40.71 & 68.69 & 12.73 & 33.33 & 52.93 & 13.74 \\
      UNO~\cite{ncd-uno} & 45.90 & 59.80 & 32.00 & 47.46 & 62.22 & \underline{32.69} & 46.47 & 57.98 & \underline{34.95} & 46.93 & 58.79 & 35.07 & 38.75 & 45.25 & \underline{32.24} \\  
      ComEx~\cite{ncd-comex} & 48.59 & 64.85 & 32.32 & 47.48 & 64.04 & 30.91 & 46.59 & 58.99 & 34.19 & 47.73 & 60.81 & 34.65 & 41.37 & 50.71 & 32.02 \\
      IIC~\cite{ncd-iic} & \underline{53.24} & 71.91 & \underline{34.55} & 51.57 & 70.51 & 32.63 & 47.73 & 60.81 & 34.65 & 49.60 & 61.82 & \underline{37.37} & 40.20 & 49.09 & 31.31 \\
      rKD~\cite{ncd-rkd} & 52.33 & \underline{72.12} & 32.53 & \underline{53.74} & \underline{75.15} & 32.32 & \underline{50.81} & 68.48 & 33.13 & \underline{52.63} & \underline{71.92} & 33.33 & \underline{46.26} & \underline{62.42} & 30.10 \\
      \midrule
      \textbf{RAPL (Ours)} & \textbf{58.99} & \textbf{79.19} & \textbf{38.79} & \textbf{58.99} & \textbf{78.59} & \textbf{39.39} & \textbf{57.07} & \textbf{71.92} & \textbf{42.22} & \textbf{61.01} & \textbf{74.75} & \textbf{47.27} & \textbf{50.20} & \textbf{64.65} & \textbf{35.76} \\
      \bottomrule
    \end{tabular}
  \end{adjustbox}
\caption{Compared results on test splits of SoyAgeing-\{R1, R3, R4, R5, R6\} with task-agnostic protocol. Rankstats IL represents the implementation of an incremental classifier. The best results are marked in bold, and the second-best results are underlined.}
\label{tab:task-agnostic}
\end{table*}

\begin{table}[t]
  \centering
  \begin{adjustbox}{width=1\linewidth}
    \begin{tabular}{cccccc}
      \toprule
      Method & R1 & R3 & R4 & R5 & R6 \\
      \midrule
      RankStats IL~\cite{ncd-rankstat} & 12.12 & 14.55 & 11.72 & 11.92 & 10.71 \\
      UNO~\cite{ncd-uno} & 31.92 & 33.37 & 34.30 & 34.91 & 31.07 \\
      ComEx~\cite{ncd-comex} & 33.13 & 32.32 & 33.94 & \underline{35.15} & \underline{33.33} \\
      IIC~\cite{ncd-iic} & \underline{34.34} & \underline{34.55} & \underline{35.71} & 34.44 & 32.07 \\
      rKD~\cite{ncd-rkd} & 33.74 & 32.73 &  35.35 & 34.14 & 32.53 \\
      \midrule
      \textbf{RAPL (Ours)} & \textbf{43.03} & \textbf{46.06} & \textbf{49.90} & \textbf{45.25} & \textbf{41.82} \\
      \bottomrule
    \end{tabular}
  \end{adjustbox}
\caption{Compared results on unlabeled train splits of SoyAgeing-\{R1, R3, R4, R5, R6\} with task-aware protocol.}
\label{tab:task-aware}
\end{table}

\subsection{Optimization Objective}
Our training process comprises a pre-train phase and a discover phase.
In the pre-train phase, we train the whole network with the proposed CRA and PSL modules on $\mathcal{D}^l$ to learn discriminative region representation and generic distribution information in embedding space:
\begin{equation}
    \mathcal{L}_{pre-train} = \alpha \mathcal{L}^{CRA} + \beta \mathcal{L}^{PC} + \gamma \mathcal{L}^{REG},
    \label{eq:pre-train}
\end{equation}
where weight parameters $\alpha$, $\beta$ and $\gamma$ control each loss term. Specifically, $\mathcal{L}^{CRA}$ encourages the network to explore discriminative local features, $\mathcal{L}^{PC}$ pull the features and their corresponding proxies close, and $\mathcal{L}^{REG}$ promotes regularized class distribution in embedding space.

In the discover phase, we initialize proxy set $\mathcal{P}^{new}$ by $k$-means~\cite{kmeans} on unlabeled data in $\mathcal{D}^{u}$. 
Note that, $\mathcal{P}^{old}$ are fixed during this phase for providing consolidated guidance to distinguish old classes and recognize new classes. 
Thus, we removed proxy regularization in $\mathcal{L}_{pretrain}$ and incorporate proxy-guided contrastive learning in discover phase:
\begin{equation}
    \mathcal{L}_{discover} = \alpha \mathcal{L}^{CRA} + \beta \mathcal{L}^{PC} + \delta \mathcal{L}^{PCL},
    \label{eq:discover}
\end{equation}
where $\delta$ is weight parameter for proxy-guided contrastive learning.
For the final label assignment, we perform $k$-means directly on the extracted feature vector $V$.




\section{Experiment}
\subsection{Experimental Setup}
\textbf{Data.} Our experiments choose five datasets from~\cite{ultra-dataset21} for UFG-NCD task, namely SoyAgeing-\{R1, R3, R4, R5, R6\}.
These datasets contain leaves from five sequential growth stages of soybeans. 
Following the protocol in~\cite{ncd-rankstat}, we take the initial 99 categories as $C^l$, and the subsequent ones as $C^u$.
For NCD scenario, each class encompasses 5 images for both training and testing. For generalized class discovery (GCD), there are 8 training images and 2 testing images per class, where unlabeled training data consists of 50\% of the images from $C^l$, in addition to all images from $C^u$.

\noindent\textbf{Evaluation Protocol.}
Following~\cite {ncd-rankstat,gcd-gcd}, 
we evaluate baselines with classification accuracy for labeled images and the average clustering accuracy (ACC) for unlabeled ones.
Given the application of k-means~\cite{kmeans} for assigning classes to all images, ACC calculations are extended across all data.
The ACC is defined as $ACC = \frac{1}{N}\sum^{N}_{i=1}\mathbbm{1}(y_i = perm_{max}(\hat y_i))$, where $N$ is the number of instances for clustering, and $perm_{max}$ is the optimal permutation derived from Hungarian 
algorithm~\cite{hungarian}.
Following~\cite{ncd-rankstat,ncd-uno,ncd-rkd}, we evaluate all methods with \textit{task-agnostic} and \textit{task-aware} protocols, respectively, indicating separate or joint assessments of instances from both old and new classes.

\noindent\textbf{Implementation Details.}
To obtain more discriminative representation, following common implementation in Ultra-FGVC~\cite{ultra-maskcov21,ultra-spare22}, we train all methods using a ResNet-50~\cite{resnet} backbone with ImageNet~\cite{imagenet} pre-trained weights, where the dimension $D=2048$ for feature maps and feature vector.
Furthermore, akin to~\cite{gcd-gcd}, we project the feature maps through a three-layer MLP projection head before applying SemiPL, and discard it at test-time.
We resize input images into 448$\times$448 and perform standard data augmentation as~\cite{ultra-cle-vit23}, generating two views for each image in the training process.
For our RAPL, we optimize the whole architecture using SGD with momentum~\cite{sgd}.
For each training phase, the initial learning rate is 0.01 which we decay with a linear warm-up and cosine annealed scheduler 
, and weight decay is set to $10^{-4}$.
Specifically, we first pre-train the network on $D^l$ for 100 epochs, and then jointly fine-tune the network on both $D^l$ and $D^u$ for another 100 epochs. 
The batch size is set to 24, temperature $\tau$ is set to 0.1 in contrastive loss, and $\beta$ is set to 2 and 1 in the pre-train and discover phase respectively. We adjust the loss weights based on ACC of ``All'' classes, as detailed in~\cref{hyperparam}.
All experiments are conducted on a single RTX 4090 GPU. 
More comprehensive details are in the Appendix. 


\subsection{Main Results}
We adopt five representative NCD methods acting as baselines of UFG-NCD, including
RankStats IL\cite{ncd-rankstat}, UNO~\cite{ncd-uno}, ComEX~\cite{ncd-comex}, IIC~\cite{ncd-iic}, and rKD~\cite{ncd-rkd}.
Results for all methods, using both task-agnostic and task-aware protocols, are detailed in~\cref{tab:task-agnostic} (test subset) and~\cref{tab:task-aware} (unlabeled training subset), respectively. 
The accuracy results in terms of ``All'', ``Old'', and ``New'', which denote all instances, and instances from old or new classes. 

Overall, as shown in~\cref{tab:task-agnostic}, our RAPL consistently outperforms all baselines by a significant margin. Specifically, our method outperforms the best results of baselines, which from state-of-the-art NCD methods, \ie, IIC\cite{ncd-iic} or rKD\cite{ncd-rkd} with a minimum gain of 3.94\% (on SoyAgeing-R6), and a maximum of 8.38\% (on SoyAgeing-R5) for ``All'' classes. 
Interestingly, while rKD shows superior accuracy for ``Old'' classes, it underperforms in learning ``New'' classes, likely due to its explicit knowledge distillation of old classes to prevent catastrophic forgetting.
Our RAPL method demonstrates consistent improvements in both ``Old'' and ``New'' categories, proving the effectiveness of our CRA strategy in capturing generic local representation for UFG images. 
This success is further bolstered by knowledge learning and transfer via SemiPL, which integrates classification knowledge within class-specified proxies and enhances feature-proxy similarity relations. 

The results under task-aware protocol in~\cref{tab:task-aware} show that our method outperforms baselines on unlabeled training images. 
In particular, our method gains more considerable improvements than in~\cref{tab:task-agnostic}, \eg, 8.49\% on SoyAgeing-R6, 14.19\% on SoyAgeing-R4. 
Note that, SoyAgeing-R6 represents the final growth cycle of the soybean, where leaves of each class differentiate into various appearances which leads to large intra-class variation and makes representation learning harder.
Thus, the results of SoyAgeing-R6 are relatively lower than those on other UFG datasets, because insufficient representation learning on labeled data leads to weak knowledge transfer on unlabeled ones. 

\begin{table}[t]
  \centering
  \begin{adjustbox}{width=1\linewidth}
    \begin{tabular}{c|ccccc|ccc}
      \toprule
      & \multicolumn{5}{c}{Component} & \multicolumn{3}{|c}{SoyAgeing-R1~\cite{ultra-dataset21}} \\
      \cmidrule(lr){2-6}\cmidrule(lr){7-9}
      & $\mathcal{L}^{PC}$ & $\mathcal{L}^{REG}$ & $\mathcal{L}^{CRA}$ & $\mathcal{L}^{PCL}$ & $\mathcal{L}^{VCL}$& All & Old & New \\
      \midrule
      (1) & \checkmark & \checkmark & \checkmark & & \checkmark & 56.36 & 74.14 & 38.59 \\
      (2) & \checkmark & \checkmark & & \checkmark & & 56.97 & 75.56 & 38.38 \\
      (3) & \checkmark & & \checkmark & \checkmark & & 58.48 & 76.97 & \textbf{40.00} \\
      (4) & \checkmark & \checkmark & \checkmark & \checkmark & & \textbf{58.99} & \textbf{79.19} & 38.79 \\
      \bottomrule
    \end{tabular}
  \end{adjustbox}
\caption{Ablation study. Each component of our method is removed in isolation, where $\mathcal{L}^{VCL}$ indicates vanilla contrastive learning without the guidance of proxies.}
\label{ablation}
\end{table}

\subsection{Ablation Study}
In~\cref{ablation}, we report the ablation results on SoyAgeing-R1 by individually removing each component of RAPL, \ie, $\mathcal{L}^{PC}$, $\mathcal{L}^{REG}$, $\mathcal{L}^{CRA}$ and $\mathcal{L}^{PCL}$, represents proxy-guided classification, proxy regularization, channel-wise region alignment, and proxy augmented contrastive learning. 

\noindent \textbf{Contrastive Learning.}
We verify the effectiveness of PCL by performing vanilla contrastive learning without the guidance of proxies. 
Comparing the Rows (1) and (4), PCL outperforms vanilla contrastive learning by 2.63\% on ``All'' classes for SoyAgeing-R1.
This implies that with the guidance of global distribution information of proxies in embedding space, the network learns a more comprehensive representation of UFG images in the discover stage.

\begin{table*}[t]
\centering
  \begin{adjustbox}{width=1\linewidth}
    \begin{tabular}{cccccccccccccccc}
      \toprule
      \multirow{2}{*}{Method} & \multicolumn{3}{c}{SoyAgeing-R1} & \multicolumn{3}{c}{SoyAgeing-R3} & \multicolumn{3}{c}{SoyAgeing-R4} & \multicolumn{3}{c}{SoyAgeing-R5} & \multicolumn{3}{c}{SoyAgeing-R6} \\
        & All & Old & New & All & Old & New & All & Old & New & All & Old & New & All & Old & New \\
      \midrule
      GCD~\cite{ncd-rankstat} & 39.65 & 49.24 & 34.85 & 38.64 & 47.73 & 34.09 & 39.90 & 45.71 & 36.99 & 40.15 & 50.00 & 35.23 & 37.79 & 46.72 & 33.33 \\
      XCon~\cite{gcd-xcon} & 33.84 & 42.93 & 29.29 & 33.08 & 42.68 & 28.28 & 35.52 & 40.15 & 33.21 & 39.98 & 49.24 & 35.35 & 34.09 & 41.92 & 30.18 \\
      \midrule
      RAPL & \textbf{52.36} & \textbf{60.89} & \textbf{48.08} & \textbf{50.87} & \textbf{58.47} & \textbf{47.07} & \textbf{53.30} & \textbf{62.10} & \textbf{48.89} & \textbf{51.14} & \textbf{61.69} & \textbf{45.86} & \textbf{44.01} & \textbf{52.42} & \textbf{39.80} \\
      \bottomrule
    \end{tabular}
  \end{adjustbox}
\caption{Compared result on train split of SoyAgeing-\{R1, R3, R4, R5, R6\} for GCD task.}
\label{tab:gcd}
\end{table*}



\noindent \textbf{Channel-wise Region Alignment.} Rows (2) and (4) show the effect of the channel-wise region alignment for UFG-NCD. 
We observe a near 2\% increase across ACC on ``All'' classes, with a more pronounced increase for the ``old'' class. 
This demonstrates that our CRA serves as a platform for effective representation learning and knowledge transfer. 
Especially when encountering labeled data, label information encourages learning discriminative features within specific common regions via $\mathcal{L}^{CRA}$. 
Besides, $\mathcal{L}^{CRA}$ facilitates representation learning on unlabeled data by sharing local knowledge across data samples and proxies. 

\noindent \textbf{Proxy Regularization.} We observe a further performance improvement across ACC on ``All'' and ``Old'' classes after introducing proxy regularization from Rows (3) and (4). 
This indicates that regularization on $\mathcal{P}^{old}$ further learns the difference between each class which is represented by proxies and embeds global distribution information with $\mathcal{P}^{old}$, which is conducive to separating old classes on embedding space with relaxed margins. 

\begin{figure}[t]
  \centering
   \includegraphics[width=1\linewidth]{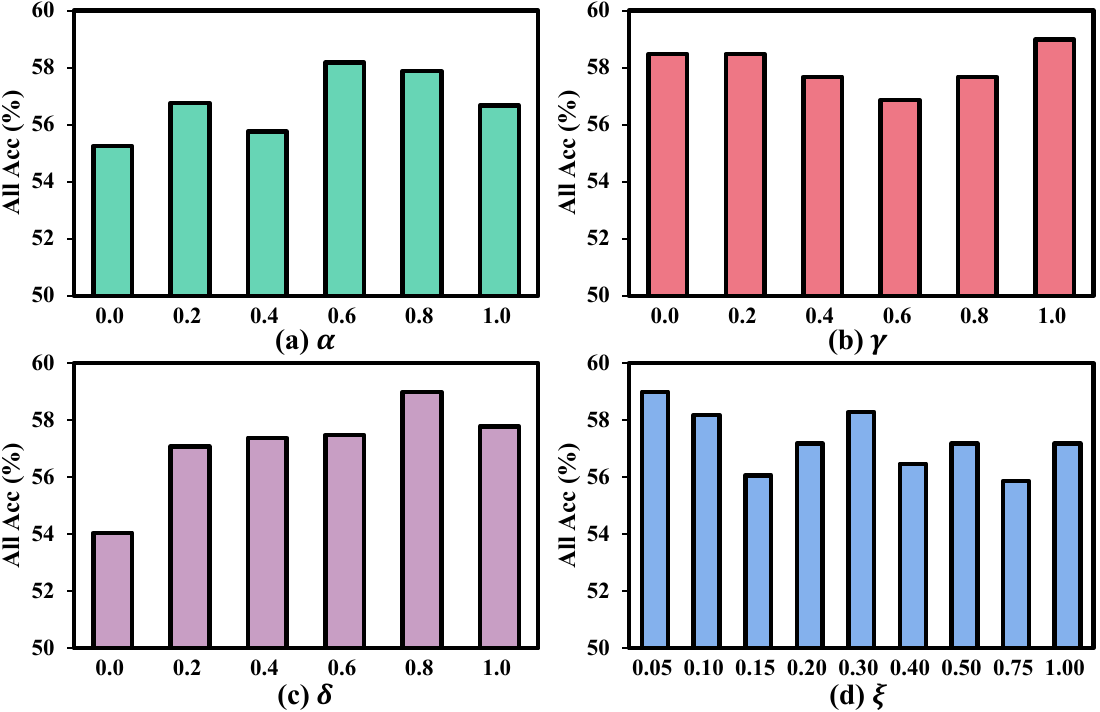}
   \caption{Impact of hyper-parameters on ACC with respect to ``All'' instances in the test dataset of SoyAgeing-R1.}
   \label{fig:hyperparam}
\end{figure}

\subsection{Hyper-Parameter Analysis}
\label{hyperparam}
In this section, we investigate the impact of the hyper-parameters in each training phase. 

\noindent \textbf{Impact of Loss Weights.} 
We study the effect of each loss weight when fixing others. 
Specifically, we first choose the optimal $\alpha$ when $\gamma$ and $\delta$ are both set to 1.0, then we determine the best $\gamma$ based on selected $\alpha$ and select the optimal $\delta$ similarly. The impact of different values of each loss weight is shown in~\cref{fig:hyperparam}(a)-(c). 
Eventually, the hyper-parameters used in our method are set to $\alpha=0.6, \gamma=1.0$, and $\delta=0.8$. 

\noindent \textbf{Impact of Number of Selected Proxies in $\mathcal{L}^{PC}$.} We study the effect of $k$ with above optimal loss weights, where we define $k=\xi \cdot C^{l}$ in $\mathcal{L}^{PC}$. As shown in~\cref{fig:hyperparam}(d), the best performance is achieved when $\xi=0.05$, indicating that focus on separating similar classes is conducive to mining 
the subtle differences between UFG images. 

\subsection{Further Investigation of RAPL}
\noindent \textbf{Evaluation on GCD Setting.}
To further evaluate the effectiveness of RAPL, we implement RAPL and two representative algorithms~\cite{gcd-gcd,gcd-xcon} on the UFG dataset under the GCD setting. 
Note that, we use ResNet-50 with ImageNet pre-trained weights for GCD and XCon for a fair comparison. 
As shown in~\cref{tab:gcd}, our RAPL consistently outperforms the GCD methods by over 10\% on SoyAgeing-\{R1,R3,R4,R5\}, 6.22\% on SoyAgeing-R6. 
Since classification heads must be trained from scratch, they are vulnerable to overfitting on the labeled classes after the pre-train phase of NCD. 
To this end, GCD and XCon turn to optimize the network with contrastive learning for more generalized feature representation in one stage.
Nevertheless, the results reproted on~\cref{tab:gcd} demonstrate that optimizing the network with RAPL boosts both representation learning on labeled and unlabeled instances, and avoids overfitting on the old data. 

\begin{figure}[t]
  \centering
   \includegraphics[width=1\linewidth]{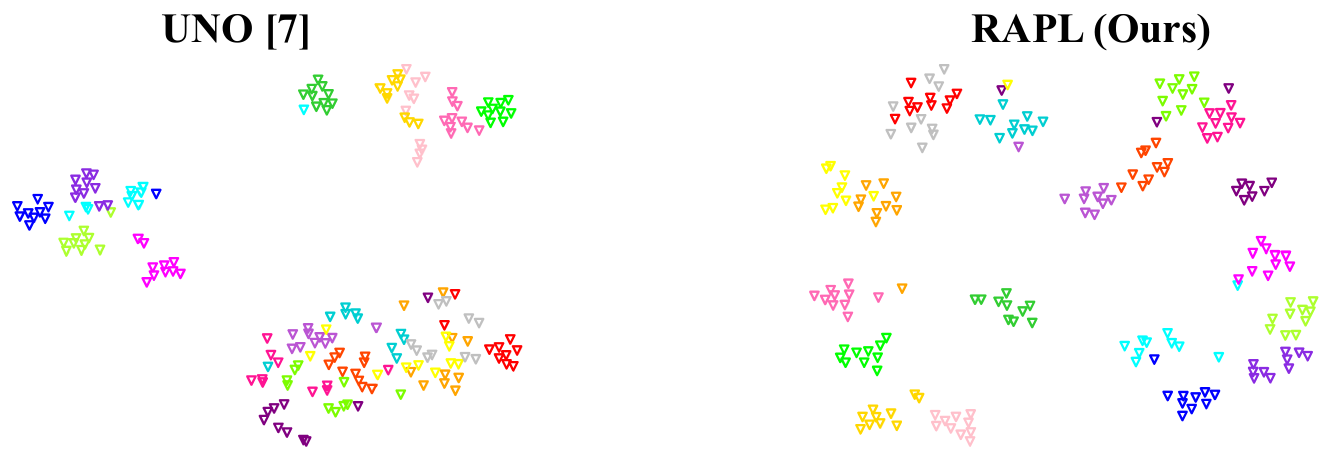}
   \caption{Visualization of features distributions of 20 unlabeled classes from SoyAgeing-R5. 
   }
   \label{fig:tsne}
\end{figure}


\noindent \textbf{Visualization of Feature Distributions.}
To qualitatively explore the clustered features on the UFG dataset, we visualize the \textit{t}-SNE embeddings based on UNO~\cite{ncd-uno} and our RAPL. Clearly, the feature embeddings optimized with our RAPL become more discriminative and less overlap in the embedding space than UNO. 

\begin{figure}[t]
  \centering
   \includegraphics[width=0.65\linewidth]{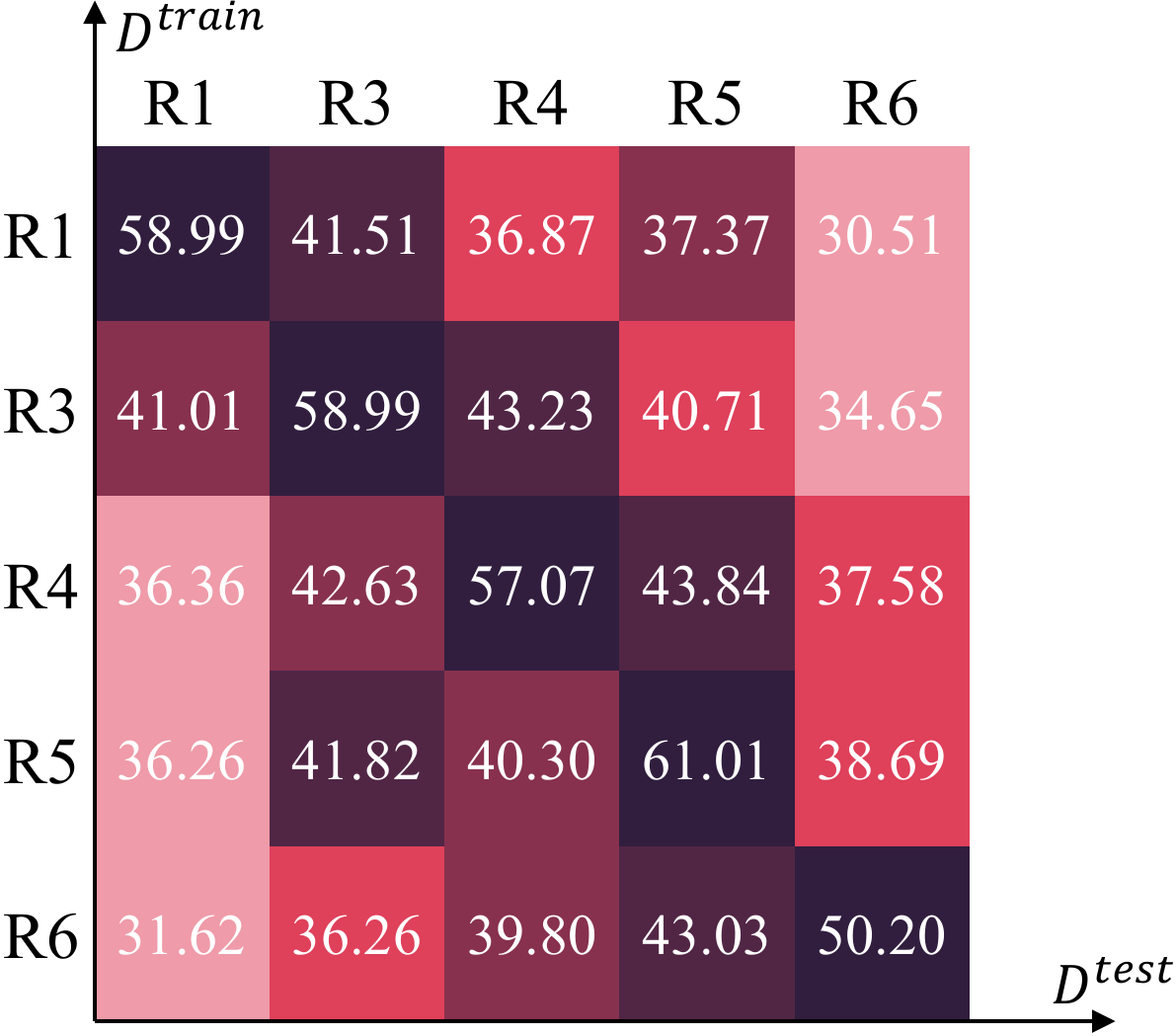}
   \caption{Generalization performance with RAPL, where $y$-axis indicates the train set and $x$-axis is the test set.}
   \label{fig:zeroshot}
\end{figure}

\noindent \textbf{Generalization Performance of RAPL.}
Since SoyAgeing-\{R1, R3, R4, R5, R6\} are collected in different growth stages of soybean, where larger numbers represent later stages, we further investigate the generalization capability of RAPL when testing on the different datasets. 
As shown in~\cref{fig:zeroshot}, RAPL achieves greater generalization capability between nearly growth stages, \eg, the performance of RAPL when training with SoyAgeing-R6 decreases on other datasets as the growth stage decreases. 
This tendency indicates that we can explore knowledge transfer across images of different ultra-fine-grained datasets to promote the classification of unlabeled UFG images.

\section{Conclusion}
We have proposed a task exploiting novel class discovery for ultra-fine-grained visual categorization, termed as UFG-NCD. 
To tackle this problem, we propose a simple yet effective framework named region-aligned proxy learning (RAPL), promoting the representation learning of local regions and the knowledge transfer from labeled to unlabeled images. 
Our RAPL is comprised of a channel-wise region alignment module to extract local representation, and a novel semi-supervised proxy learning framework to facilitate representation learning based on the relations between class-level proxies and instances. 
Experimental results on five datasets show that RAPL achieves state-of-the-art performance. In the future, RAPL has the potential to be applied to other ultra-fine-grained scenarios. 

\par\noindent\textbf{Acknowledgement.}
This work was supported by the
NSF of China under Grant Nos. 62102061, 62272083 and 62306059, and by the Liaoning Provincial NSF under Grant 2022-MS-137 and 2022-MS-128. Besides, this work was partially supported by the EU Horizon projects ELIAS (No. 101120237) and AI4Trust (No. 101070190).
We also acknowledge the support of the MUR PNRR project \textbf{iNEST}-Interconnected Nord-Est Innovation Ecosystem (ECS00000043) funded by the NextGenerationEU.
{
    \small
    \bibliographystyle{ieeenat_fullname}
    \bibliography{main}
}
\end{document}